\documentclass[runningheads]{llncs}

\usepackage{graphicx}
\usepackage{multirow}
\usepackage{tabularx}
\usepackage{amssymb}
\usepackage{amsmath}
\usepackage{booktabs}
\usepackage{mathtools}
\usepackage{pifont}
\usepackage{bm}
\usepackage{hyperref}
\usepackage{cite}

\usepackage{comment}

\begin{document}

\title{Hierarchical Self-Supervised Learning for Medical Image Segmentation Based on Multi-Domain Data Aggregation}
\titlerunning{HSSL}

\author{Hao Zheng
\and
Jun Han \and
Hongxiao Wang \and 
Lin Yang \and \\
Zhuo Zhao \and 
Chaoli Wang \and 
Danny Z. Chen}
\authorrunning{H. Zheng et al}
\institute{Department of Computer Science and Engineering, University of Notre Dame,\\ Notre Dame, IN 46556, USA \\
\email{hzheng3@nd.edu}  }

\maketitle
\begin{abstract}
A large labeled dataset is a key to the success of supervised deep learning, but for medical image segmentation, it is highly challenging to obtain sufficient annotated images for model training. In many scenarios, unannotated images are abundant and easy to acquire. 
Self-supervised learning (SSL) has shown great potentials in exploiting raw data information and representation learning. 
In this paper, we propose Hierarchical Self-Supervised Learning (HSSL), a new self-supervised framework that boosts medical image segmentation by making good use of unannotated data. 
Unlike the current literature on task-specific self-supervised pretraining followed by supervised fine-tuning, we utilize SSL to learn task-agnostic knowledge from heterogeneous data for various medical image segmentation tasks. 
Specifically, we first aggregate a dataset from several medical challenges, then pre-train the network in a self-supervised manner, and finally fine-tune on labeled data. We develop a new loss function by combining contrastive loss and classification loss, and pre-train an encoder-decoder architecture for segmentation tasks. 
Our extensive experiments show that multi-domain joint pre-training benefits downstream segmentation tasks and outperforms single-domain pre-training significantly.
Compared to learning from scratch, our method yields better performance on various tasks (e.g., $+0.69\%$ to $+18.60\%$ in Dice with $5\%$ of annotated data). With limited amounts of training data, our method can substantially bridge the performance gap with respect to denser annotations (e.g., $10\%$ vs.~$100\%$ annotations). 
\end{abstract}

\section{Introduction}\label{Sec-Intro}

\begin{figure}[t]
    \centering
    \includegraphics[width=\columnwidth]{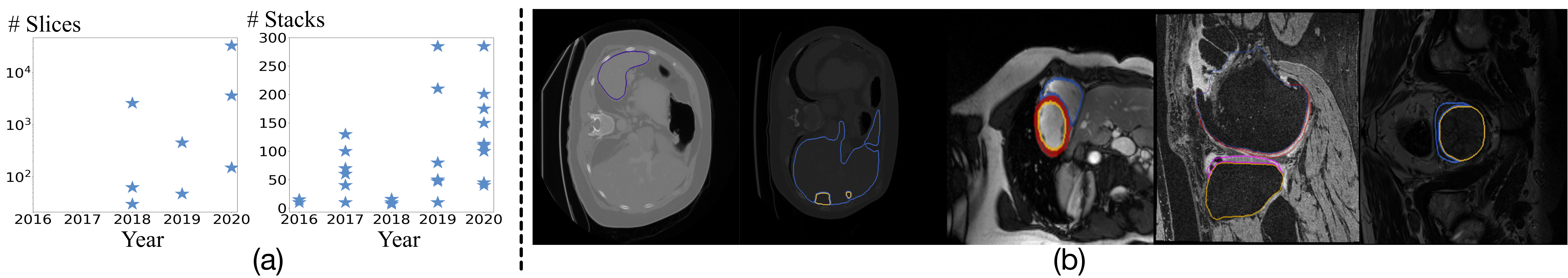}
    \caption{(a) The number of images for each medical image segmentation challenge every year since 2016 at MICCAI (left: 2D images; right: 3D stacks). (b) Diverse medical image and mask examples: spleen, liver \& tumours, cardiovascular structures, knee bones \& cartilages, and prostate. }
\label{fig:example}
\end{figure}

Although supervised deep learning has achieved great success on medical image segmentation~\cite{ronneberger2015u,liang2019cascade,zheng2019hfa,huang2020unet}, it heavily relies on sufficient good-quality manual annotations which are usually hard to obtain due to expensive acquisition, data privacy, etc. Public medical image datasets are normally smaller than the generic image datasets (see Fig.~\ref{fig:example}(a)), and may hinder improving segmentation performance. 
Deficiency of annotated data has driven studies to explore alternative solutions. 
Transfer learning fine-tunes models pre-trained on ImageNet for target tasks~\cite{zhou2017fine,he2019rethinking,zhou2019models}, but it could be impractical and inefficient due to the pre-defined model architectures~\cite{liu2019anisotropic} and is not as good as transferred from medical images due to image characteristics differences~\cite{zhou2019models}.
Semi-supervised learning utilizes unlimited amounts of unlabeled data to boost performance, but it usually assumes that the labeled data sufficiently covers the data distribution, and needs to address consequent non-trivial challenges such as adversarial learning~\cite{zhang2017deep,madani2018semi} and noisy labels~\cite{yu2019uncertainty,zheng2020Anno}. 
Active learning selects the most representative samples for annotation~\cite{yang2017suggestive,zhou2017fine,zheng2019biomedical} but focuses on saving manual effort and does not utilize unannotated data. 
Considering these limitations and the fact that considerable unlabeled medical images are easy to acquire and free to use, we seek to answer the question: \textit{Can we improve segmentation performance with limited training data by directly exploiting raw data information and representation learning?}

Recently, self-supervised learning (SSL) approaches, which initialize models by constructing and training surrogate tasks with unlabeled data, attracted much attention due to soaring performance on representation learning~\cite{doersch2015unsupervised,noroozi2016unsupervised,larsson2016learning,pathak2016context,gidaris2018unsupervised,oord2018representation,hjelm2018learning,grill2020bootstrap} and downstream tasks~\cite{zhou2019models,chen2019self,zhuang2019self,ouyang2020self,tao2020revisiting,chaitanya2020contrastive}. 
It was shown that the learned representation by \emph{contrastive learning}, a variant of SSL, gradually approaches the effectiveness of representations learned through strong supervision, even under circumstances when only limited data or a small-scale dataset is available~\cite{chen2020simple,he2020momentum}. 
However, three key factors of contrastive learning have not been well explored for medical segmentation tasks: 
(1) A medical image dataset is often insufficiently large due to the intrusive nature of some imaging techniques or expensive annotations (e.g., 3D(+T) images), which suppresses self-supervised pre-training and hinders representation learning using a single dataset. 
(2) The contrastive strategy considers only congenetic image pairs generated by different transformations used in data augmentation, which suppresses the model from learning task-agnostic representations from heterogeneous data collected from different sources (see Fig.~\ref{fig:example}(b)). 
(3) Most studies focused on extracting high-level representations by pre-training the encoder while neglecting to learn low-level features explicitly and initialize the decoder, which hinders the performance of dense prediction tasks such as semantic segmentation.

To address these challenges, in this paper, we propose a new \emph{hierarchical self-supervised learning} (HSSL) framework to pre-train on heterogeneous unannotated data and obtain an initialization beneficial for training multiple downstream medical image segmentation tasks with limited annotations. 
First, we investigate available public challenge datasets on medical image segmentation and propose to aggregate a multi-domain (modalities, organs, or facilities) dataset. In this way, our collected dataset is considerably larger than a task-specific dataset and the pretext model is forced to learn task-agnostic knowledge (e.g., texture, intensity distribution, etc). 
Second, we construct pretext tasks at multiple abstraction levels to learn hierarchical features and explicitly force the model to learn richer semantic features for segmentation tasks on medical images. Specifically, our HSSL utilizes contrasting and classification strategies to supervise image-, task-, and group-level pretext tasks. We also extract multi-level features from the network encoding path to bridge the gap between low-level texture and high-level semantic representations. 
Third, we attach a lightweight decoder to the encoder and pre-train the encoder-decoder architecture to obtain a suitable initialization for downstream segmentation tasks.

We experiment on our aggregated dataset composed of eight medical image segmentation tasks and show that our HSSL is effective in utilizing multi-domain data to initialize model parameters for target tasks and achieves considerably better segmentation, especially when only limited annotations are available.

\section{Methodology}\label{sec:Method}

\begin{figure}[!t]
    \centering
    \includegraphics[width=0.95\columnwidth]{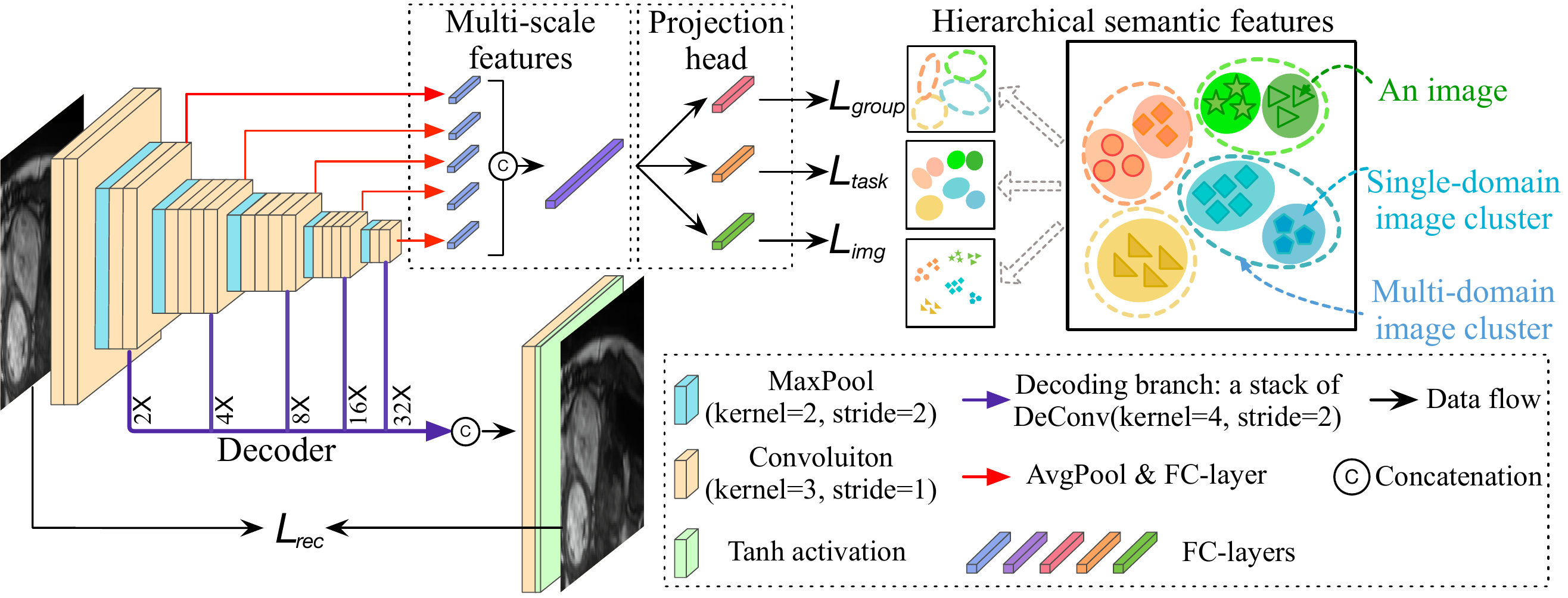}
    \caption{An overview of our proposed hierarchical self-supervised learning (HSSL) framework (best viewed in color). The backbone encoder builds a pyramid of multi-scale features from the input image, forming a rich latent vector. Then it is stratified to represent hierarchical semantic features of the aggregated multi-domain data, supervised by different pretext tasks in the hierarchy. Besides, an auxiliary reconstruction pretext task helps initialize the decoder.} 
\label{fig:overview}
\end{figure}

We discuss the necessity and feasibility of aggregating multi-domain image data and show how to construct such a dataset in Sect.~\ref{sec:M_agg}, and then introduce our hierarchical self-supervised learning pretext tasks (shown in Fig.~\ref{fig:overview}) in Sect.~\ref{sec:M_hssl}. After pre-training, we fine-tune the trained encoder-decoder network on downstream segmentation tasks with limited annotations.

\subsection{Multi-Domain Data Aggregation }\label{sec:M_agg}

\noindent
{\bf Necessity.} As shown in Fig.~\ref{fig:example}(a), most publicly available medical image segmentation datasets are of relatively small sizes. Yet, recent progresses on contrastive learning empirically showed that training on a larger dataset often learns better representations and brings larger performance improvement in downstream tasks~\cite{chen2020simple,chen2020big,he2020momentum}. Similarly, a larger dataset is beneficial for supervised classification tasks and unsupervised image reconstruction tasks, because such a dataset tends to be more diverse and better cover the true image space distribution. 

\noindent
{\bf Feasibility.} 
First, there are quite a few medical image dataset archives (e.g., \href{https://www.cancerimagingarchive.net/}{TCIA}) and public challenges (e.g., \href{https://grand-challenge.org/challenges/}{Grand Challenge}). Typical imaging modalities (CT, MRI, X-ray, etc) of multiple regions-of-interest (ROIs, organs, structures, etc) are covered. 
Second, common/similar textures or intensity distributions are shared among different datasets (see Fig.~\ref{fig:example}(b)), and their raw images may cover the same physical regions (e.g., abdominal CT for the spleen dataset and liver dataset). 
Therefore, an aggregated multi-domain dataset can (1) enlarge the data size of a shared image space and (2) force the model to distinguish different contents from the raw images. In this way, task-agnostic knowledge is extracted.

\noindent
{\bf Dataset Aggregation.} 
To ensure the effectiveness of multi-domain data aggregation, three principles should be considered. 
(1) Representativeness: The datasets considered for aggregation should cover a moderate range of medical imaging techniques/modalities. 
(2) Relevance: The datasets considered should not drastically differ in content/appearance. Otherwise, it is easy for the model to distinguish them and a less common feature space is shared among them.  
(3) Diversity: The datasets considered should benefit a range of applications. 
In this work, we focus on CT and MRI of various ROIs (i.e., heart, liver, prostate, pancreas, knee, and spleen). 
The details of aggregated dataset are shown in Table~\ref{tab:dataset}. 

\subsection{Hierarchical Self-Supervised Learning (HSSL)} \label{sec:M_hssl}

Having aggregated multiple datasets, $\mathcal{D}=\{D_1, D_2, \ldots$, $D_N\}$, where $D_i$ is a dataset for a certain segmentation task.
A straightforward method to use $\mathcal{D}$ is to directly extend some known pretext tasks (e.g., SimCLR~\cite{chen2020simple}) and conduct joint pre-training.
However, such pretext tasks only explicitly force the model to learn a global representation and are not tailored for the target segmentation tasks. Hence, taking imaging techniques and prior knowledge (e.g., appearance, ROIs) into account, we propose to extract richer semantic features from hierarchical abstract levels and devise the network for target segmentation tasks. 

\begin{figure}[t]
    \centering
    \includegraphics[width=0.9\columnwidth]{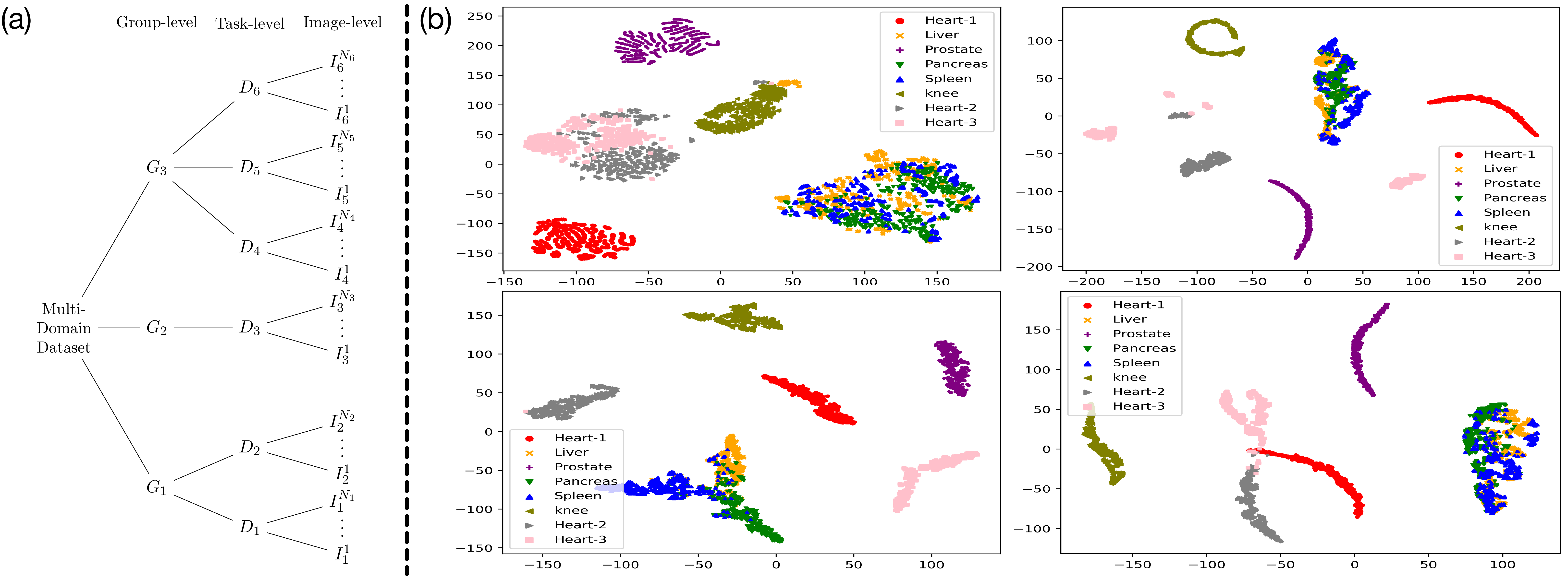}
    \caption{(a) An example of the hierarchical structure of a multi-domain dataset. Each chosen dataset/task $D_i$ forms a domain consisting of a set of images $\{I_i^k\}_{k=1}^{N_i}$, where $N_i$ is the total number of images in $D_i$. Multiple tasks form a multi-domain cluster called a \emph{group} ($G_j$). 
    (b) t-SNE projection~\cite{maaten2008visualizing} of extracted features (best viewed in color). 
    Top-left: $F_{VGG-19}$; top-right: $F_{image}$; bottom-left: $F_{task}$ (forming single-domain task-level clusters as in Table~\ref{tab:dataset}); bottom-right: $F_{group}$ (forming multi-domain group-level clusters as in Table~\ref{tab:dataset}). }
    \label{fig:tree_tsne}
\end{figure}

We formulate three hierarchical levels (see Fig.~\ref{fig:tree_tsne}(a)).
(1) \emph{Image-level}: Each image $I$ is a learning subject; we want to extract distinguishable features of $I$ w.r.t.~another image, regardless of which dataset it originally comes from or what ROIs it contains. Specifically, we follow the state-of-the-art SimCLR~\cite{chen2020simple} and build positive and negative pairs with various data augmentations.   
(2) \emph{Task-level}: Each $D_i$ is originally imaged for a specific purpose (e.g., CT for spleen). Generally, images belonging to a same dataset are similar inherently. As shown in Fig.~\ref{fig:tree_tsne}(b), images of different modalities and ROIs are easier to distinguish. For abdominal CTs of spleen and liver, although the images are similar, their contents are different. Thus, each task's dataset forms a single domain of certain ROI and image types. 
(3) \emph{Group-level}: 
Despite the differences among different segmentation tasks, the contents of images may show a different degree of similarity. For example, in the physical space, liver CT scans have overlapping with spleen CT scans; cardiac MRIs scanned for different purposes (e.g., diverse cardiovascular structures) contain the same ROI (i.e., the heart) regardless of the image size and contrast. In this way, we categorize multiple domains of images into a group, which forms a multi-domain cluster in the feature space.
Assigned with both task-level and group-level labels, each image constitutes a tuple $(I, y^t, y^g)$, where $t$ and $g$ are task-class and group-class, respectively (see Table~\ref{tab:dataset}). 

Further, to better aggregate low- and high-level features from the encoder, we compress multi-scale feature vectors from the feature pyramid and concatenate them together, and then attach three different projection heads to automatically extract hierarchical representations (see Fig.~\ref{fig:overview}).

\noindent
{\bf Image-Level Loss.} 
Given an input image $I$, the contrastive loss is formulated as: $
l(\tilde{I}, \hat{I}) 
= -\log\frac{\exp\{{sim(\tilde{z}, \hat{z})/{\tau}}\}}{\exp\{{sim(\tilde{z}, \hat{z})/{\tau}}\} + \sum_{\bar{I}\in \Lambda^-}\exp\{{sim(\tilde{z}, \bar{z})/{\tau}}\} }$, 
where $\tilde{z}=P_l(E(\tilde{I}))$,  $\hat{z}=P_l(E(\hat{I}))$, $\bar{z}=P_l(E(\bar{I}))$, $P_l(\cdot)$ is the image-level projection head, $E(\cdot)$ is the encoder, $\tilde{I}$ and $\hat{I}$ are two different augmentations of image $I$ (i.e., $\tilde{I}=\tilde{t}(I)$ and $\hat{I}=\hat{t}(I)$), $\bar{I} \in \Lambda^-$ consisting of all negative samples of $I$,
and $\tilde{t}$, $\hat{t} \in \mathcal{T}$ are two augmentations. The augmentations $\mathcal{T}$ include random cropping, resizing, blurring, and adding noise. $sim(\cdot,\cdot)$ is cosine similarity, and 
$\tau$ is a temperature scaling parameter. 
Given our multi-domain dataset $\mathcal{D}$, the image-level loss is defined as: $\mathcal{L}_{img} = \frac{1}{|\Lambda^{+}|} \sum_{\forall(\tilde{I}, \hat{I})\in \Lambda^{+}}{[l(\tilde{I}, \hat{I})+l(\hat{I},\tilde{I}) ]}$, where $\Lambda^+$ is a set of all similar pairs sampled from $\mathcal{D}$. 

\noindent
{\bf Task-Level Loss \& Group-Level Loss.}
Given task-class and group-class, we formulate task- and group-level pretext tasks as classification tasks. The training objectives are: $\mathcal{L}_{task} = -\sum_{c=1}^T y_c^t \log(p_c^t)$; $\mathcal{L}_{group} = -\sum_{c=1}^G y_c^g \log(p_c^g)$, where $p_c^t=P_t(E(I))$, $p_c^g=P_g(E(I))$, $P_t(\cdot)$ (or $P_g(\cdot)$) is the task-level (or group-level) projection head, $E(\cdot)$ is the encoder, $y_c^t$ (or $y_c^g$) is the task-class (or group-class) of input image $I$, and $T$ (or $G$) is the number of classes of tasks (or groups).  

We visualize some sample learned features in Fig.~\ref{fig:tree_tsne}(b), in which the hierarchical layout is as expected, implying that our model is capable of extracting richer semantic features at different abstract levels of the input images.

\noindent
{\bf Decoder Initialization.}
A decoder is also indispensable for semantic segmentation tasks. We devise a multi-scale decoder and formulate a reconstruction pretext task. The loss is defined as: $\mathcal{L}_{rec} = \frac{1}{|\mathcal{D}|}\sum_{I \in \mathcal{D}} || S(E(I)) - I ||_{2}$, where $E(\cdot)$ is the encoder, $S(\cdot)$ is the decoder, and $||\cdot||_{2}$ is the $L_{2}$ norm.

In summary, we combine the hierarchical self-supervised losses at all the levels and the auxiliary reconstruction loss to jointly optimize the model: $\mathcal{L}_{total} = \lambda_1\mathcal{L}_{img} + \lambda_2\mathcal{L}_{task} + \lambda_3\mathcal{L}_{group} + \lambda_4\mathcal{L}_{rec}$, where $\lambda_i (i=1,2,3,4)$ are the weights to balance loss terms. For simplicity, we let $\lambda_1 = \lambda_2= \lambda_3 = 1/3, \lambda_4 = 50$.

\noindent
{\bf Segmentation.}
Once trained, the encoder-decoder can be fine-tuned for downstream multi-domain segmentation tasks. For a give task $D_i$, we acquire some annotations (e.g., $10\%$) and optimize the network with cross-entropy loss.

\section{Experiments and Results}\label{sec:Exp}

\begin{table}[t]
\centering
\caption{Details of our data obtained from public sources. The left two columns: their task-classes and group-classes based on our multi-domain data aggregation principles. }
\label{tab:dataset}
\scriptsize
\resizebox{\columnwidth}{!} 
{
\begin{tabular}{|c|c|c|c|c|c|}
\hline
Task ID & Group ID & ROI-Type &  Segmentation class & \# of slices & Source \\ \hline
 1   & 1 & Heart-MRI    & 1: left atrium    & 1262  & LASC~\cite{tobon2015benchmark} \\ \hline
 2   & 2 & Liver-CT     & 1: liver, 2: tumor   & 4342  & LiTS~\cite{bilic2019liver} \\ \hline
 3   & 3 & Prostate-MRI & 1: central gland, 2: peripheral zone  & 483  & MSD~\cite{simpson2019large} \\ \hline
 4   & 2 & Pancreas-CT  & 1: Pancreas, 2: tumor  & 8607 & MSD~\cite{simpson2019large} \\ \hline
 5   & 2 & Spleen-CT    & 1: spleen   & 1466 & MSD~\cite{simpson2019large} \\ \hline
\multirow{ 2}{*}{ 6}   & \multirow{ 2}{*}{4}  & \multirow{ 2}{*}{Knee-MRI}     & \multirow{ 2}{*}{ \shortstack{1: femur bone, 2: tibia bone, \\ 3: femur cartilage, 4: tibia cartilage} }   & \multirow{ 2}{*}{ 8187 } & \multirow{ 2}{*}{Knee~\cite{yin2010logismos}} \\
 & & & & & \\ \hline
 7   & 1 & Heart-MRI    & \multirow{ 2}{*}{\shortstack{ 1: left ventricle, 2: right ventricle, \\ 3: myocardium }} & 1891 & ACDC~\cite{bernard2018deep} \\ \cline{1-3}\cline{5-6}
 8   & 1 & Heart-MRI    &   & 3120 & M\&Ms~\cite{MMM20} \\ \hline
\end{tabular}
}
\end{table}

\noindent{\bf Datasets and Experimental Setup.} 
We employ multiple MRI and CT image sets from 8 different data sources with distribution shift, and sample 2D slices from each stack (see Table \ref{tab:dataset} for a summary of their sample numbers and downstream tasks). 
Each dataset is split into $X_{tr}$, $X_{val}$, and $X_{te}$ in the ratios of $7:1:2$. We use all images for the pre-training stage and then fine-tune the pre-trained network with labeled images from $X_{tr}$. We experiment with different amounts of training data $X_{tr}^{s}$, where $s\in\{5\%,10\%, 100\%\}$ denotes the ratio of $\frac{X_{tr}^{s}}{X_{tr}}$.
The segmentation accuracy is measured by the Dice-Sørensen Coefficient.

\noindent{\bf Implementation Details.} 
For self-supervised pre-training, we use ResNet-34~\cite{he2016deep} as the base encoder network, FC layers to obtain latent vectors, and sequential DeConv layers to reconstruct images. Detailed structures can be found in Supplementary Material. The model is optimized using Adam with a linear learning rate scaling for 1$k$ epochs (initial learning rate = $3e^{-4}$). 
For segmentation tasks, we optimize the network using Adam with the ``poly" learning rate policy, $L_r\times \big(1-\frac{epoch}{\# epoch}\big)^{0.9}$, where the initial learning rate $L_r=5e^{-4}$ and $\# epoch=10k$. Random cropping and rotation are applied for augmentation.
In all the experiments, the mini-batch size is 30 and input image size is $192\times192$.

\begin{table}[!t]
\centering
\caption{Quantitative results on Task-1 (heart), Task-3 (prostate), and Task-5 (spleen). Dice scores for each class are listed and the average scores are in parentheses. TFS: training from scratch. Same network architecture is used for fair comparison in all the experiments. Our HSSL achieves the best performance in most settings (in bold). }
\label{tab:mainres}
\resizebox{\columnwidth}{!}
{
\begin{tabular}{c|c| c|c|c|c|c|c}
\toprule
 Task-\# & Anno. & TFS & Rotation~\cite{gidaris2018unsupervised} & In-painting~\cite{pathak2016context} & MoCo~\cite{he2020momentum} & SimCLR~\cite{chen2020simple} &  HSSL (Ours) \\ \hline 
\multirow{ 3}{*}{ 1 } & $5\%$ & 71.56   & 72.83 & 65.40 & 75.97 & 73.45 & {\bf 81.46} \\  
 & $10\%$ & 79.64 & {\bf 82.31} & 81.99 & 79.07 & 81.19 & 81.79 \\  
 & $100\%$ & 85.81 & 87.43 & 86.56 & 87.19 & 87.06 & {\bf 87.65} \\  \hline
\multirow{ 3}{*}{ 3 } & $5\%$ & 20.65; 47.56 (34.10) & 28.74; 67.11 (47.93) & 20.13; 52.16 (36.14) & 29.55; 64.95 (47.25) & 39.67; 68.35 ({\bf 54.01}) & 35.30; 70.08 (52.69) \\  
 & $10\%$ & 40.10; 66.95 (53.53) & 44.15; 70.63 (57.39) & 33.81;67.14 (50.48) & 40.16; 67.98 (54.07) & 46.04; 70.39 (58.22) & 46.97; 72.21 ({\bf 59.59}) \\  
 & $100\%$ & 50.19; 76.74 (63.47) & 55.21; 78.21 (66.71) & 53.19 77.97 (65.59) & 56.31; 77.59 (66.95) & 56.53; 77.86 (67.20) & 58.80; 78.35 ({\bf 68.58}) \\  \hline
 \multirow{ 3}{*}{ 5 } & $5\%$ & 48.75 & 56.74 & 47.86 & 54.91 & 63.40 & {\bf 67.35} \\  
 & $10\%$ & 67.44 & 74.68 & 71.30 & 68.22 & 78.25 & {\bf 80.95} \\  
 & $100\%$ & 85.88 & 86.96 & 85.96 & 85.75 & 87.76 & {\bf 88.45} \\  \bottomrule
\end{tabular}
}
\end{table}

\begin{figure}[!t]
    \centering 
    \includegraphics[width=\linewidth]{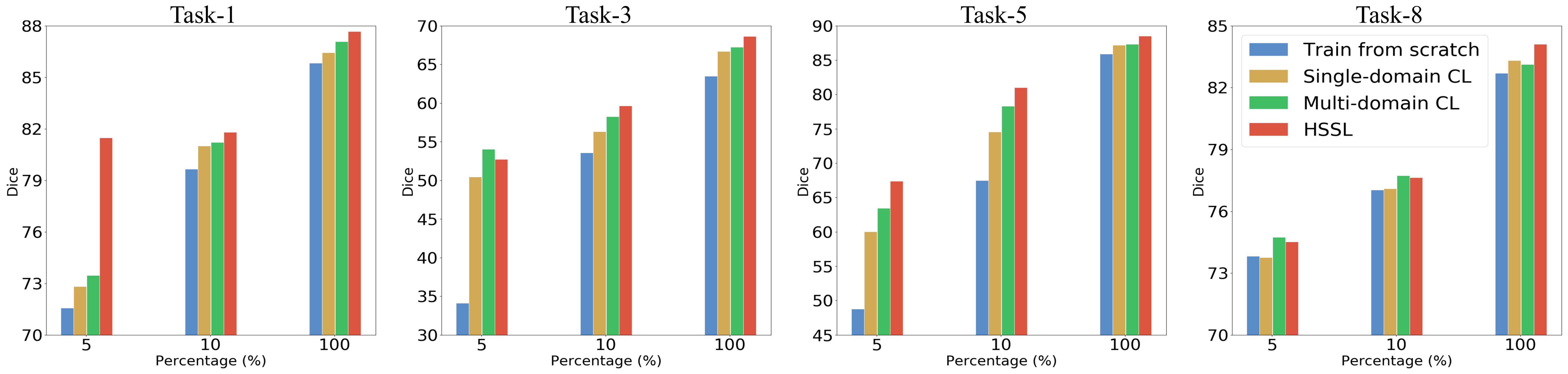} 
    \caption{Quantitative results of TFS \emph{vs.}~single-domain CL \emph{vs.}~multi-domain CL \emph{vs.}~HSSL for Task-1/-3/-5/-8 with different ratios ($5\%$, $10\%$, $100\%$) of labeled data, respectively.} 
    \label{fig:res}
\end{figure}

\begin{figure}[!t]
    \centering 
    \includegraphics[width=0.9\columnwidth]{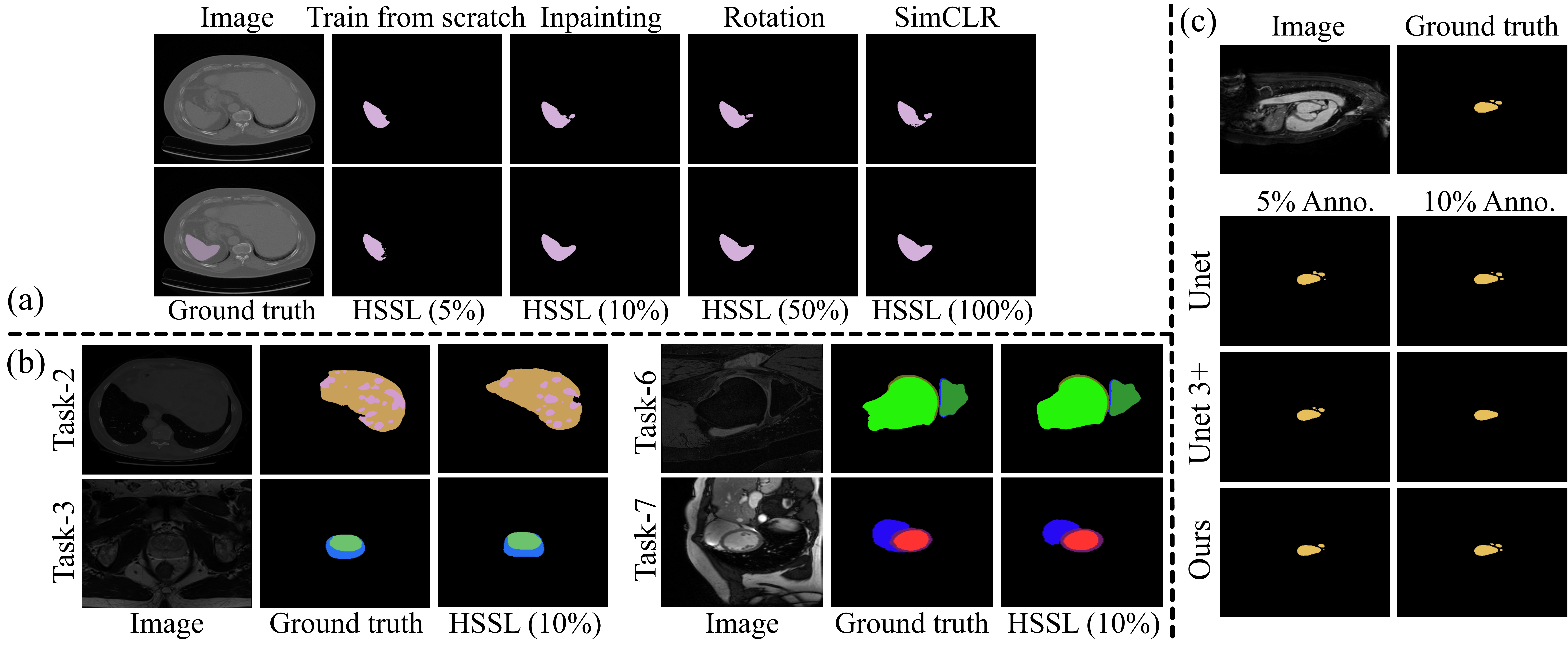}
    \caption{Qualitative comparison (best viewed in color). (a) Top: results of different methods on Task-5 ($10\%$ annotated data); Bottom: results of our HSSL with different ratios of annotated data. (b) Results of Task-2/-3/-6/-7 ($10\%$ annotated data). (c) Results of different models on Task-1 trained with $5\%$ and $10\%$ annotated data, respectively.} 
    \label{fig:QualRes}
\end{figure}

\noindent{\bf Main Results. }
Our approach contributes to the ``pre-training + fine-tuning" scheme in two aspects: hierarchical self-supervised learning (HSSL) and multi-domain data aggregation. 
{\bf (1)} \emph{Effectiveness of HSSL.} We compare with state-of-the-art pretext task training methods~\cite{pathak2016context,gidaris2018unsupervised,chen2020simple,he2020momentum} on seven downstream segmentation tasks, and quantitative results of three representative tasks are summarized in Table~\ref{tab:mainres}. 
First, our method surpasses training from scratch (TFS) substantially, showing the effectiveness of better model initialization. More can be found in Supplementary Material. 
Second, our approach outperforms known SSL-based methods in almost all the settings, indicating a better capability to extract features for segmentation tasks. 
Third, our HSSL can more effectively boost performance, especially when extremely limited annotations are available (e.g., $+18.60\%$ with $5\%$ annotated data on Task-3), implying potential applicability when abundant images are acquired but few are labeled. 
Fourth, with more annotations, our method can further improve accuracy and achieve state-of-the-art performance (e.g., $+1.84\%$ to $+2.57\%$ with $100\%$ annotated data over~TFS). 
Qualitative results are given in Fig.~\ref{fig:QualRes} and Supplementary Material.
{\bf (2)} \emph{Effectiveness of Multi-Domain Data Aggregation.} We conduct pre-training on single-domain and aggregated multi-domain data, and compare the segmentation performances. ``Single-domain CL" and ``Multi-domain CL" are all based on the state-of-the-art SimCLR~\cite{chen2020simple}. 
As sketched in Fig.~\ref{fig:res}, one can see that multi-domain data aggregation consistently outperforms (sometimes significantly) single-domain pre-training (e.g., with $10\%$ annotated data on Task-5, multi-domain CL and HSSL outperform single-domain CL by $3.74\%$ and $6.41\%$, respectively). 
This suggests that more data varieties can provide complementary information and help improve the overall performance.

\begin{table}[!t]
    \centering
    \begin{tabular}{c c}
    \begin{minipage}{.64\linewidth}
      \centering
      \caption{Quantitative results of different models on Task-1/-3/-5 with $5\%$, $10\%$, and $50\%$ annotated data, respectively. Our HSSL achieves the best performance in most the settings (highest scores in bold).}
      \label{tab:sota}
\resizebox{\columnwidth}{!}
{
\begin{tabular}{c|c| c|c|c| c|c|c| c|c|c}
\hline
\multirow{ 2}{*}{Method} & \multirow{ 2}{*}{ \shortstack{Param.\\(M)}} &  \multicolumn{3}{c|}{Task-1 (heart)} & \multicolumn{3}{c|}{Task-3 (prostate)} & \multicolumn{3}{c}{Task-5 (spleen)} \\ \cline{3-11}
 & & $5\%$ & $10\%$ & $50\%$ & $5\%$ & $10\%$ & $50\%$ & $5\%$ & $10\%$ & $50\%$  \\ \hline
UNet~\cite{ronneberger2015u} & 39.40 & 75.43 & 77.72 & 86.75 & 38.19 & 49.44 & 62.61 & 54.71 & 62.81 & 81.48 \\ \hline
UNet3+~\cite{huang2020unet} & 26.97 & 78.48 & 78.81 & {\bf 87.52} & 42.06 & 50.94 & 63.50 & 60.05 & 64.83 & 82.74 \\ \hline
HSSL (Ours) & 22.07 & {\bf 81.46} & {\bf 81.79} & 87.02 & {\bf 52.69} & {\bf 59.59} & {\bf 66.64} & {\bf 67.35} & {\bf 80.95} & {\bf 85.86} \\ 
\hline
\end{tabular}
}
    \end{minipage} & \hspace{.1cm}
    \begin{minipage}{.34\linewidth}
      \centering
      \caption{Ablation study of loss functions on Task-1 \& Task-5 w/ 5\% anno. data.}
      \label{tab:ablation}
\resizebox{\columnwidth}{!}
{
\begin{tabular}{cccc|c|c}
\hline
$L_{rec}$ & $L_{img}$ & $L_{task}$ & $L_{group}$ & Task-1 & Task-5 \\\hline
 \checkmark &  &  &  & 65.71 & 46.13 \\ 
  & \checkmark &  &  & 73.45 & 63.40 \\ 
 \checkmark & \checkmark &  &  & 77.26 & 65.01 \\ 
 \checkmark & \checkmark & \checkmark &  & 79.32 & 66.67 \\ 
 \checkmark & \checkmark & \checkmark & \checkmark & 81.46 & 67.35 \\ 
\hline
\end{tabular}
}
    \end{minipage}
    \end{tabular}
\end{table}

\noindent
{\bf Discussions. }
{\bf (1)} \emph{Comparison with State-of-the-Art Models.} 
As shown in Table~\ref{tab:sota}, our method outperforms the state-of-the-art UNet3+~\cite{huang2020unet} significantly in almost all the settings. Further, with limited annotated data (e.g., $5\%$), our method bridges the performance gap significantly w.r.t. the results obtained by training with more annotated data. 
Also, our model is most lightweight, and thus efficient as well. Qualitative results are given in Fig.~\ref{fig:QualRes}(c).
{\bf (2)} \emph{Ablation Study.} As shown in Table~\ref{tab:ablation}, each hierarchical loss contributes to representation learning and leads to segmentation improvement.

\section{Conclusions}
In this paper, we proposed \emph{hierarchical self-supervised learning}, a novel self-supervised framework that learns hierarchical (image-, task-, and group-levels) and multi-scale semantic features from aggregated multi-domain medical image data. A decoder is also initialized for downstream segmentation tasks. Extensive experiments demonstrated that joint training on multi-domain data by our method outperforms training from scratch and conventional pre-training strategies, especially in limited annotation scenarios.

\subsubsection{Acknowledgement.} This research was supported in part by the U.S. National Science Foundation through grants IIS-1455886, CCF-1617735, CNS-1629914, and IIS-1955395.

\bibliographystyle{splncs04}
\bibliography{reference}

\end{document}


\begin{center}
\textbf{\LARGE Supplementary Materials}
\end{center}

\begin{table}
    \centering
    \caption{The configurations of Encoder, Decoder, Feature Fusion, \& Projection Head.}
    \label{tab:network}
    \begin{tabular}{c c}
    \begin{minipage}{.45\linewidth}
      \centering
\resizebox{\columnwidth}{!}
{
\begin{tabular}{|c |c| c|}
    \hline
    \multicolumn{3}{|c|}{Encoder}  \\  \hline
    {Layer name} & {Operators} & {Output size} \\ \hline
    Input  &  $-$  &  192$\times$192$\times$3 \\ \hline
    conv1-1  &  3$\times$3, 64 &  192$\times$192$\times$64   \\ 
    conv1-2  &  3$\times$3, 64, /2 &  96$\times$96$\times$64   \\ \hline 
    layer-1 &   $\begin{Bmatrix}
                3\times3, 64\\
                3\times3, 64
              \end{Bmatrix} \times 3$ & 48$\times$48$\times$64 \\   \hline
    layer-2 &   $\begin{Bmatrix}
                3\times3, 128\\
                3\times3, 128
              \end{Bmatrix} \times 4$ & 24$\times$24$\times$128 \\   \hline
    layer-3 &   $\begin{Bmatrix}
                3\times3, 256\\
                3\times3, 256
              \end{Bmatrix} \times 6$ & 12$\times$12$\times$256 \\   \hline
    layer-4 &   $\begin{Bmatrix}
                3\times3, 512\\
                3\times3, 512
              \end{Bmatrix} \times 3$ & 6$\times$6$\times$512 \\   \hline
\end{tabular}
}
    \end{minipage} & \hspace{.1cm}
    \begin{minipage}{.4\linewidth}
      \centering
\resizebox{\columnwidth}{!}
{
\begin{tabular}{|c|c|c|c|}
    \hline
    \multicolumn{4}{|c|}{Decoder}  \\  \hline
    Previous layer & Layer name & Operators & Output size  \\ \hline
    \multirow{5}{*}{layer-4} & \multirow{5}{*}{up-4} & $4\times4, 64, 2\times$ & 12$\times$12$\times$64 \\
     & & $4\times4, 32, 2\times$ & 24$\times$24$\times$32 \\
     & & $4\times4, 16, 2\times$ & 48$\times$48$\times$16 \\ 
     & & $4\times4, 8, 2\times$ & 96$\times$96$\times$8 \\
     & & $4\times4, 4, 2\times$ & 192$\times$192$\times$4 \\
    \hline
    \multirow{4}{*}{layer-3} & \multirow{4}{*}{up-3} & $4\times4, 32, 2\times$ & 24$\times$24$\times$32 \\
     & & $4\times4, 16, 2\times$ & 48$\times$48$\times$16 \\ 
     & & $4\times4, 8, 2\times$ & 96$\times$96$\times$8 \\
     & & $4\times4, 4, 2\times$ & 192$\times$192$\times$4 \\
    \hline
    \multirow{3}{*}{layer-2} & \multirow{3}{*}{up-2} & $4\times4, 16, 2\times$ & 48$\times$48$\times$16 \\ 
     & & $4\times4, 8, 2\times$ & 96$\times$96$\times$8 \\
     & & $4\times4, 4, 2\times$ & 192$\times$192$\times$4 \\
    \hline
    \multirow{2}{*}{layer-1} & \multirow{2}{*}{up-1} & $4\times4, 8, 2\times$ & 96$\times$96$\times$8 \\
     & & $4\times4, 4, 2\times$ & 192$\times$192$\times$4 \\
    \hline
    up-1$\sim$up-4 & Fout-0 & concatenation & 192$\times$192$\times$16 \\
    \hline
    Fout-0 & Fout-1 & $4\times4, 16$ & 192$\times$192$\times$16 \\
    \hline
    Fout-1 & Fout-2 & $4\times4, $ N & 192$\times$192$\times$N \\
    \hline
\end{tabular}
}
    \end{minipage} \\ 
    \begin{minipage}{.5\linewidth}
      \centering
\resizebox{\columnwidth}{!}
{
\begin{tabular}{|c|c|c|c|}
    \hline
    \multicolumn{4}{|c|}{Feature Fusion}  \\  \hline
    Previous layer & Layer name & Operators & Output size  \\ \hline
    layer-4 & V4 & Avg-Pool(6), 2 $fc-$layers & 512 \\
    layer-3 & V3 & Avg-Pool(12), 2 $fc-$layers & 256 \\
    layer-2 & V2 & Avg-Pool(24), 2 $fc-$layers & 128 \\
    V2$\sim$V4 & V$_{all}$ & concatenation & 896 \\
    \hline
\end{tabular}
}
    \end{minipage} & \hspace{.1cm}
    \begin{minipage}{.35\linewidth}
      \centering
\resizebox{\columnwidth}{!}
{
\begin{tabular}{|c|c|c|c|}
    \hline
    \multicolumn{4}{|c|}{Projection Head}  \\  \hline
    Previous layer & Layer name & Operators & Output size  \\ \hline
    V$_{all}$ & H1 & 2 $fc-$layers & 512 \\
    V$_{all}$ & H2 & 2 $fc-$layers & 512 \\
    V$_{all}$ & H3 (i.e., $F_I$) & 2 $fc-$layers & 512 \\ \hline
    H1 & $F_G$ & 2 $fc-$layers & 4 \\
    H2 & $F_T$ & 2 $fc-$layers & 8 \\
    \hline
\end{tabular}
}
    \end{minipage}
    \end{tabular}
\end{table}

\begin{figure}
\begin{center}
   \includegraphics[width=\columnwidth]{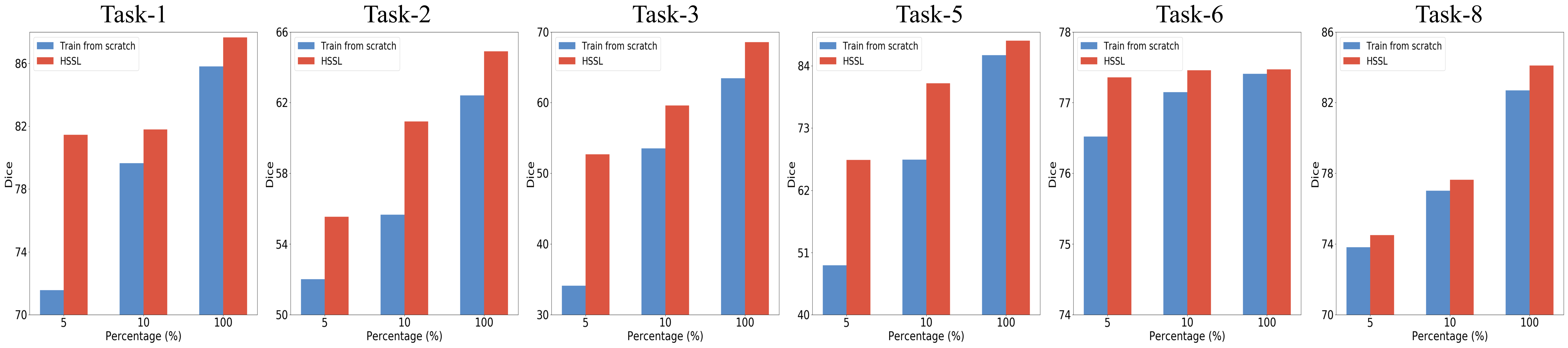}
\end{center}
   \caption{Quantitative results of Task-1 (heart), Task-2 (liver), Task-3 (prostate), Task-5 (spleen), Task-6 (knee), and Task-8 (heart) with $5\%$, $10\%$, and $100\%$ annotated data, respectively. }
\label{fig:results}
\end{figure}

\begin{figure}[h]
    \centering 
    \includegraphics[width=\columnwidth]{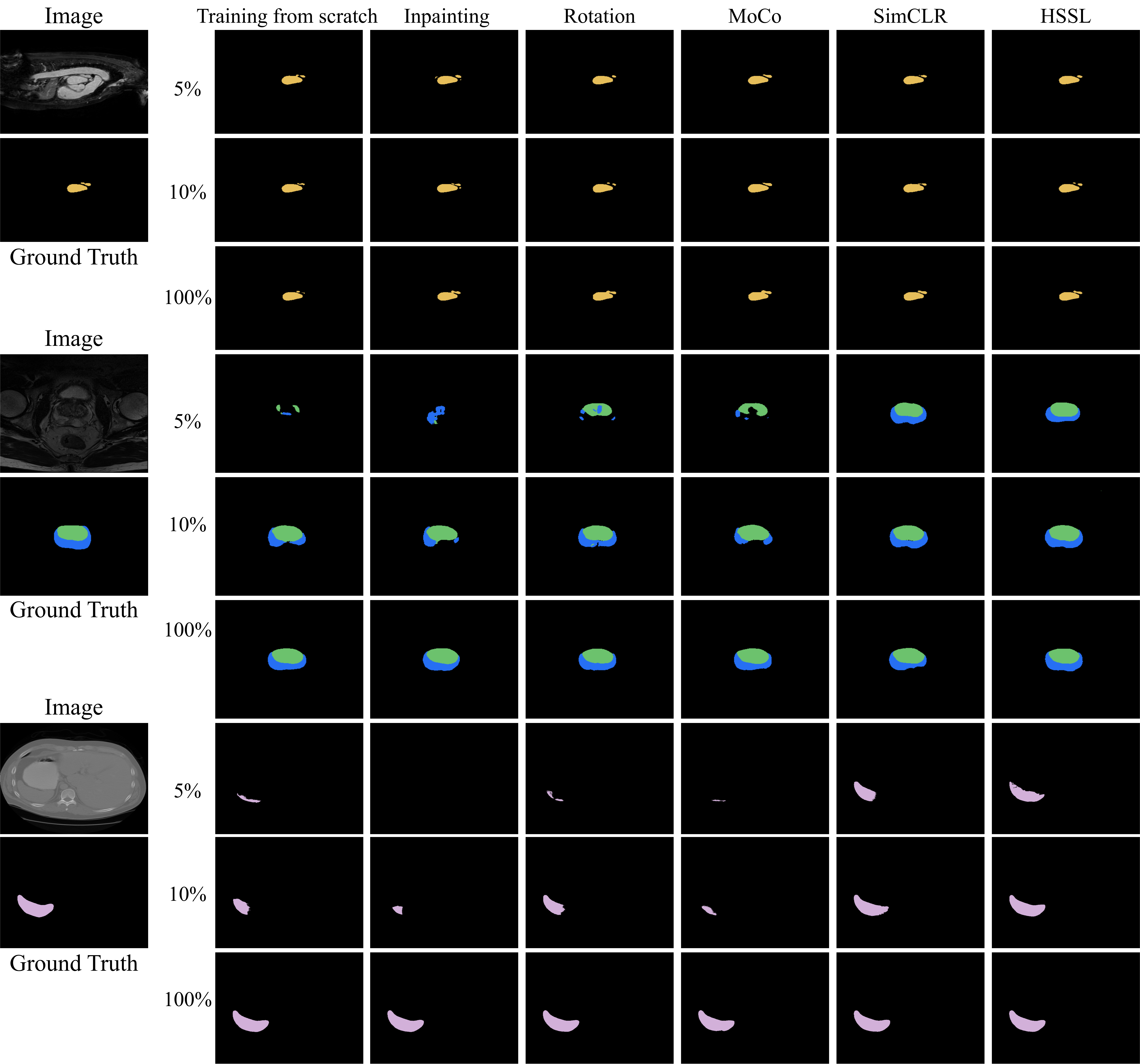}
    \caption{Qualitative results of several known methods and our HSSL on Task-1 (heart), Task-3 (prostate), and Task-5 (spleen) trained with different ratios ($5\%$, $10\%$, and $100\%$) of annotated data, respectively. One can observe that our HSSL consistently yields good results, especially in extremely sparse annotation scenarios. 
    } 
    \label{fig:VisRes1}
\end{figure}

\begin{figure}[h]
    \centering 
    \includegraphics[width=\columnwidth]{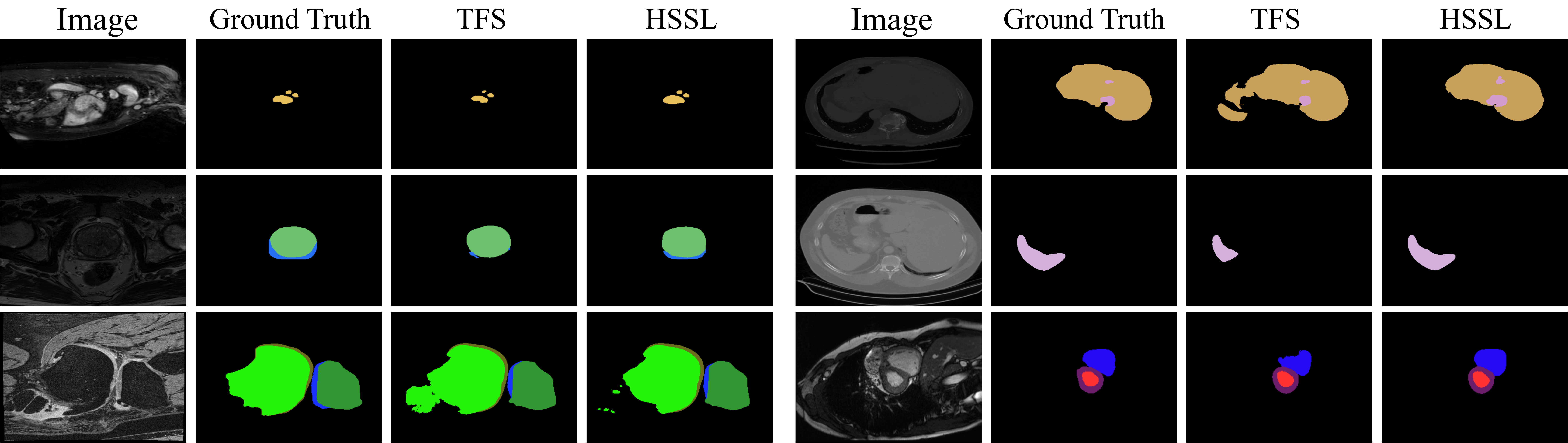}
    \caption{Comparison of qualitative results between TFS (training from scratch) and our proposed HSSL using $10\%$ annotated data on six segmentation tasks: Task-1 (heart), Task-2 (liver), Task-3 (prostate), Task-5 (spleen), Task-6 (knee), and Task-7 (heart) (from left to right, top to bottom). 
    } 
    \label{fig:VisRes2}
\end{figure}